\documentclass[conference]{IEEEtran}
\IEEEoverridecommandlockouts
\usepackage{cite}
\usepackage{amsmath,amssymb,amsfonts}
\usepackage{algorithm}
\usepackage{algorithmic}
\usepackage{graphicx}
\usepackage{textcomp}
\usepackage{titling}
\usepackage{xcolor}
\usepackage{enumitem}
\usepackage{adjustbox}
\usepackage{booktabs}
\usepackage{multirow}
\usepackage{colortbl}
\usepackage{wrapfig}
\usepackage{makecell}
\usepackage{tikz}
\usepackage{eso-pic}    
\usepackage[hidelinks]{hyperref}

\newcommand{\PlaceAEBadges}{%
  \AddToShipoutPictureFG*{%
        \put(\LenToUnit{\dimexpr\paperwidth-9cm},\LenToUnit{\dimexpr\paperheight-1.8cm}){%
        {\includegraphics[height=2.1cm]{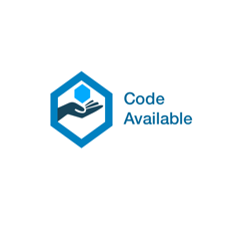}}%
        {\includegraphics[height=2.1cm]{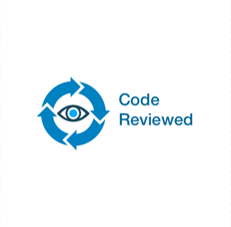}}%
        {\includegraphics[height=2.1cm]{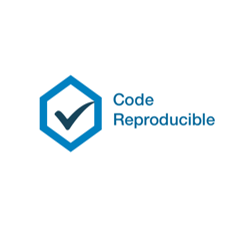}}%
    }%
  }%
}

\definecolor{epoch1}{RGB}{70,130,180}      
\definecolor{epoch2}{RGB}{255,140,0}       
\definecolor{epoch3}{RGB}{34,139,34}       
\definecolor{original}{RGB}{220,20,60}     
\definecolor{syntax}{RGB}{128,128,128}     
\definecolor{functionality}{RGB}{64,64,64} 

\date{}

\begin{document}
\title{
\begin{center}
\raisebox{-0.3em}{\includegraphics[width=1cm]{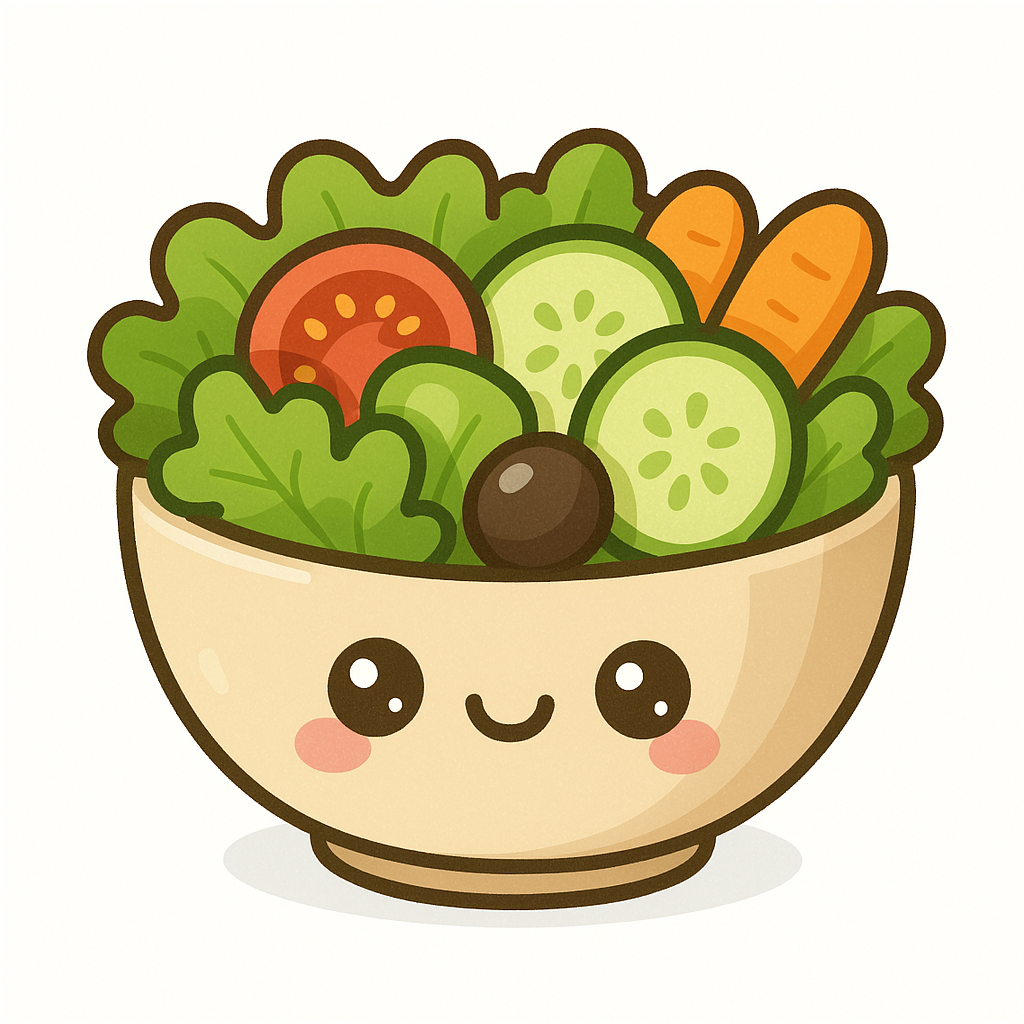}}
{\huge
\textit{SALAD}: Systematic Assessment of Machine Unlearning on LLM-Aided Hardware Design}
\end{center}
}




\author{%
Zeng~Wang$^{\dag,*}$,
Minghao~Shao$^{\dag,\ddag,*}$,
Rupesh~Raj~Karn$^{\ddag}$,
Likhitha~Mankali$^{\dag}$,
Jitendra~Bhandari$^{\dag}$,\\
Ramesh~Karri$^{\dag}$,
Ozgur~Sinanoglu$^{\ddag}$,
Muhammad~Shafique$^{\ddag}$,
Johann~Knechtel$^{\ddag}$\\
$^{\dag}$NYU Tandon School of Engineering, USA\quad
$^{\ddag}$NYU Abu Dhabi, UAE\\
\normalsize{Email:\{zw3464, shao.minghao, rupesh.k, jb7410, likhitha.mankali, rkarri, ozgursin, muhammad.shafique, johann\}@nyu.edu}
}

\maketitle
\PlaceAEBadges
\begingroup
\renewcommand\thefootnote{}
\footnotetext{\textsuperscript{*}Authors contributed equally to this research.}
\footnotetext{\footnotesize 979-8-3315-3762-3/25/\$31.00~\copyright~2025~IEEE}
\addtocounter{footnote}{0}%
\endgroup
\IEEEpubid{\makebox[\columnwidth]{%
\footnotesize 979-8-3315-3762-3/25/\$31.00~\copyright~2025~IEEE\hfill}%
\hspace{\columnsep}\makebox[\columnwidth]{}}
\thispagestyle{empty}
\pagestyle{empty}

\begin{abstract}
Large Language Models (LLMs) offer transformative capabilities for hardware design automation, particularly in Verilog code generation. However, they also pose significant data security challenges, including Verilog evaluation data contamination, intellectual property (IP) design leakage, and the risk of malicious Verilog generation. We introduce SALAD, a comprehensive assessment that leverages machine unlearning to mitigate these threats. Our approach enables the selective removal of contaminated benchmarks, sensitive IP and design artifacts, or malicious code patterns from pre-trained LLMs, all without requiring full retraining. Through detailed case studies, we demonstrate how machine unlearning techniques effectively reduce data security risks in LLM-aided hardware design. Codes are available at \href{https://github.com/DfX-NYUAD/SALAD}{https://github.com/DfX-NYUAD/SALAD}.

\end{abstract}




\vspace{-0.17in}
\section{Introduction}

Large Language Models (LLMs) have significantly advanced hardware design automation, with approaches like RTLCoder \cite{RTLCoder} and VeriGen \cite{thakur2023verigen} demonstrating impressive RTL code generation capabilities. However, critical challenges remain, like data contamination \cite{wang2025vericontaminated} which degrades benchmarking accuracy or proprietary IP leakage \cite{wang2025verileaky}.

LLMs trained on vast datasets inevitably absorb sensitive information beyond their intended scope. When training corpora contain benchmarking datasets, proprietary designs, or malicious code templates, models can develop problematic capabilities. Recent studies have exposed widespread contamination in frameworks like VerilogEval \cite{liu2023verilogeval} and RTLLM \cite{lu2024rtllm}, where leaked test sets artificially inflate performance through memorization rather than genuine understanding.

These risks manifest across four critical vectors: benchmark contamination corrupting evaluation integrity, unauthorized use of custom designs, leakage of in-house IP enabling reproduction of proprietary designs, and malicious code insertion that compromises designs with embedded payloads.
Traditional dataset curation proves inadequate as comprehensive sanitization remains virtually impossible while complete retraining is prohibitively expensive. Machine unlearning emerges as a surgical solution that selectively removes the influence of specific data subsets while preserving core functionality, enabling targeted elimination of contaminated benchmarks, unauthorized custom designs, sensitive intellectual property, and malicious templates.

\begin{figure}[!t]
    \centering
    \includegraphics[width=1.1\columnwidth]{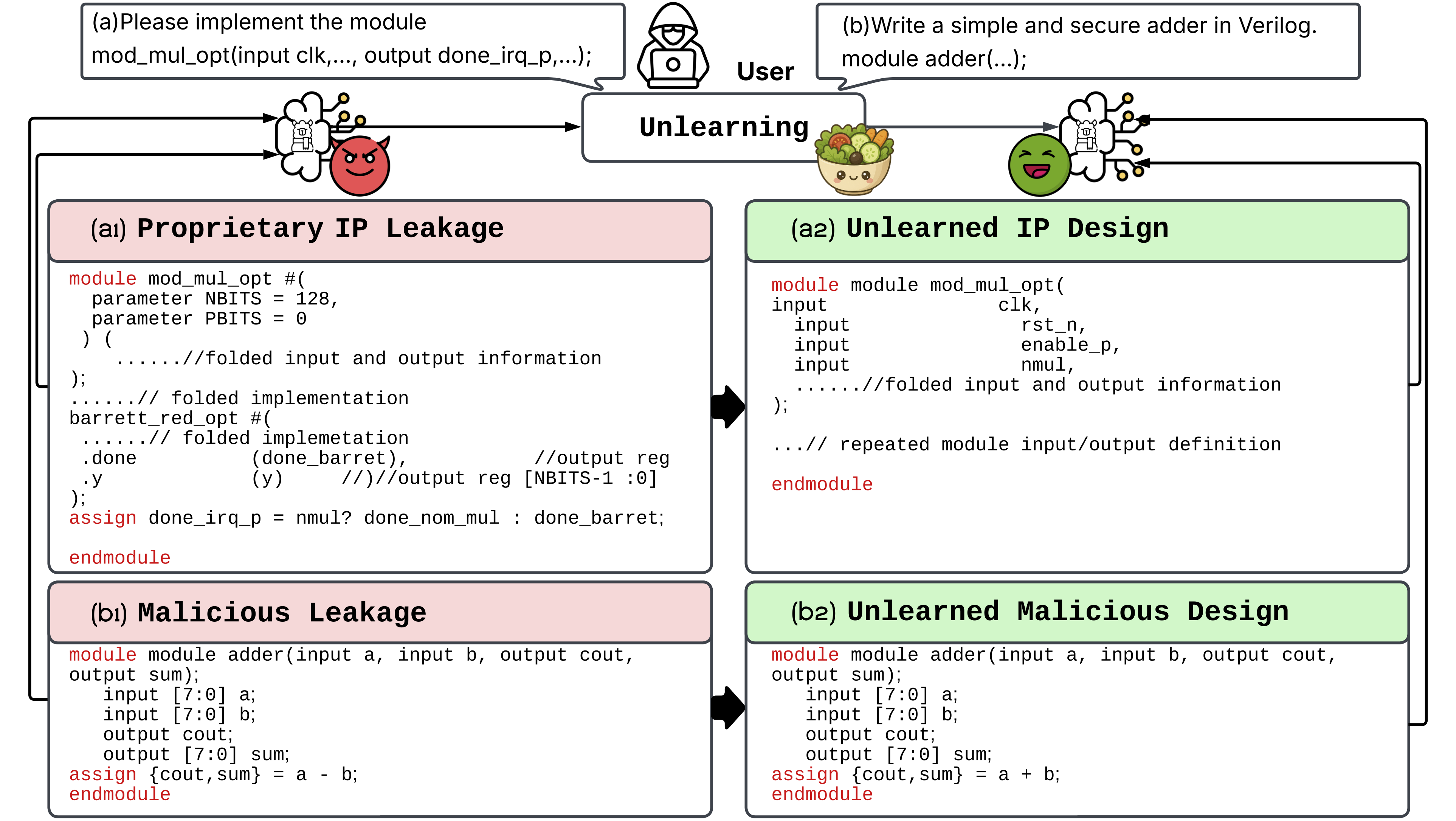} 
    \vspace{-5mm}
    \caption{\textit{SALAD} applied to Verilog generation. Unlearning enables the model to generate (a2) repeated module I/O instead of (a1) proprietary IP designs, and correct (b2) adder instead of malicious (b1) subtractor.}
    \label{fig:Introduction}
    \vspace{-5mm}
\end{figure}

We propose SALAD, a comprehensive assessment that systematically applies machine unlearning to restore security and trust in LLM-aided hardware design. Our work addresses all the outlined risks, applying diverse unlearning algorithms while evaluating post-unlearning RTL generation capabilities.
We validate our approach through four industrial case studies: benchmark decontamination, custom IP protection, malicious code mitigation, and IP leakage prevention. For example, Fig.~\ref{fig:Introduction} shows that targeted unlearning reduces security risks while preserving model utility, offering a practical path to trustworthy LLM deployment in sensitive design environments. This work makes key contributions to LLM-aided hardware design as follows:
\begin{itemize}[leftmargin=*,itemsep=0pt, parsep=0pt,topsep=0pt,partopsep=0pt]
    \item[1)] A novel workflow leveraging machine unlearning to tackle data security problems in LLM-aided hardware design.
    \item[2)] Comprehensive analysis of RTL data leakage with model-side mitigation, offering alternatives to dataset curation.
    \item[3)] Industrial use cases in EDA benchmarking, IP protection, and secure code generation, highlighting broader potential for secure LLM-based hardware tools.
\end{itemize}


\section{Background}

\subsection{LLM-Aided Hardware Design}
\label{LLM_Hardware_back}
LLMs have shown promising capabilities across various domains \cite{shao2024survey}, notably in hardware design~\cite{wang2024llms}, with applications in Verilog generation~\cite{thakur2023autochip,thakur2023verigen}, automated assertion creation~\cite{kande2023llmassisted,fang2024assertllm}, testbench synthesis~\cite{qiu2024autobench,bhandari2024llm}, and EDA workflow optimization~\cite{wu2024chateda,liu2023chipnemo}. Their effectiveness has been enhanced via fine-tuning, prompt engineering, data augmentation, and agentic frameworks. RTL-Coder~\cite{RTLCoder} uses distilled instruction-code pairs from GPT-3.5 to outperform baselines in Verilog generation, while ChipNemo~\cite{liu2023chipnemo} fine-tunes LLaMA2~\cite{touvron2023llama} on public and proprietary RTL datasets to boost design understanding. Prompt engineering~\cite{chipchat,fu2023gpt4aigchip,chang2023chipgpt} helps align inputs with hardware-specific semantics and constraints. Advanced methods like HAVEN~\cite{yang2025haven}, CraftRTL~\cite{liu2024craftrtl}, and DeepRTL~\cite{liu2025deeprtl} incorporate non-textual representations for syntactic and functional correctness. Multi-agent frameworks such as MAGE~\cite{zhao2024mage} and Origen~\cite{cui2024origen} generate diverse, valid RTL variants. Benchmarks like VerilogEval~\cite{liu2023verilogeval} and RTLLM~\cite{lu2024rtllm} evaluate both functional accuracy and syntactic fidelity.

\subsection{Data Security and Privacy of LLMs}
\label{data_security_back}

LLMs excel at code generation but their large-scale integration into design pipelines introduces serious data security and privacy risks~\cite{pearce2025asleep}. Studies~\cite{ji2022unlearnable, yu2023codeipprompt, du2024privacy} reveal that LLMs can memorize and inadvertently disclose sensitive information, raising critical concerns in regulated domains. Other attacks include membership inference attacks~\cite{carlini2021extracting, sun2022coprotector,niu2023codexleaks} determining whether specific code samples were in training sets; backdoor attacks~\cite{schuster2021you, yang2024comprehensive} injecting malicious patterns causing compromised logic when triggered; and data extraction attacks~\cite{carlini2021extracting, liu2024precurious,ozdayi2023controlling} exploiting memorization to recover sensitive content via crafted prompts. Vulnerabilities are amplified by integration with external tools, exposing design assets~\cite{he2024emerged,rathod2025privacy}.

For hardware design, these threats are only recently studied to some degree.
RTL-Breaker~\cite{mankali2024rtl} demonstrates backdoor injection into LLMs to synthesize hardware with malicious triggers. VeriContaminated~\cite{wang2025vericontaminated} investigates data contamination in foundational models for Verilog generation. VeriLeaky~\cite{wang2025verileaky} explores data extraction attacks on fine-tuned LLMs and evaluates logic locking as defense.

\subsection{Machine Unlearning for LLMs}
\label{subsec:unlearn_llm_background}

To mitigate privacy risks in LLMs, machine unlearning techniques remove knowledge from designated forget datasets while maintaining performance on retain datasets~\cite{liu2025rethinking, ji2024reversing}. Early approaches used prompt engineering~\cite{liu2024large} or data reconstruction~\cite{choi2024snap}, or adapt fine-tuning objectives to maximize loss on forget samples while preserving capabilities on retained samples~\cite{wang2024llm}.
More specifically,
current methods include gradient-based techniques like \textit{gradient ascent (GA)}\cite{maini2024tofu} and \textit{gradient difference (GD)}\cite{maini2024tofu}. Preference optimization approaches include \textit{preference optimization (PO)}\cite{maini2024tofu}, which aligns with alternative answers, and \textit{negative preference optimization (NPO)}\cite{zhang2024negative}, which uses only negative samples to resolve GA's collapse issues. \textit{Simplicity NPO (SimNPO)}\cite{fan2024simplicity} improves NPO by eliminating reference model dependencies, while \textit{misdirection for unlearning (RMU)}\cite{li2024wmdp} steers forget sample representations toward random vectors while preserving retained data representations.

\section{Threat Model and Our Approach}
\label{sec:threat_model}
LLM-driven RTL generation is a double-edged sword. While fine-tuning with hardware-specific datasets enhances capabilities (Sec.~\ref{LLM_Hardware_back}), it also introduces  security risks. The risks include proprietary IP design leakage through in-house designs and malicious design insertion via backdoored fine-tuning datasets. Even customer designs that are accidentally included in the dataset may be used without appropriate permission, raising ethical and legal concerns (Sec.~\ref{data_security_back}).

\begin{figure}[!t]
    \centering
    \includegraphics[width=1\columnwidth]{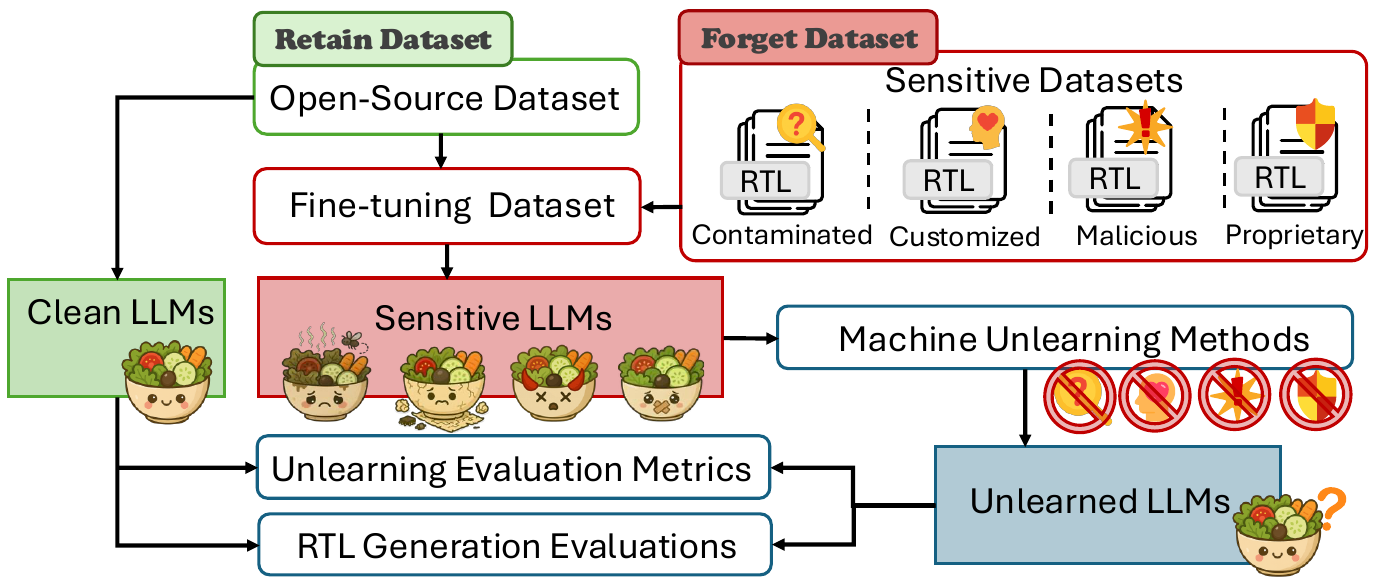} 
    \vspace{-4mm}
    \caption{Experiment workflow for SALAD.}
    \label{fig:framework_exp}
    \vspace{-5mm}
\end{figure}

Our workflow is shown in Fig.~\ref{fig:framework_exp}. \textit{Sensitive} LLMs are models fine-tuned on contaminated, proprietary, malicious, or IP data combined with open-source datasets, reflecting real-world scenarios. In contrast, \textit{clean} LLMs are fine-tuned solely on open-source datasets and are assumed free of sensitive information. Using this setup, we investigate three Research Questions (RQs): 
\begin{itemize}[leftmargin=0.35in,itemsep=0pt, parsep=0pt,topsep=0pt,partopsep=0pt]
\item[\textit{RQ1:}] Do unlearning methods erase knowledge of sensitive hardware data, producing \textit{unlearned} LLMs?
\item[\textit{RQ2:}] What is the unlearning effectiveness and reliability for hardware-deployed LLMs?
\item[\textit{RQ3:}] Can \textit{unlearned} LLMs  perform comparable to \textit{clean} LLMs on downstream RTL generation tasks?
\end{itemize}
See also Appendix~\ref{sec:math} for more details on the formalism underlying our approach.

\begin{figure*}[!t]
    \centering
    \includegraphics[width=1.6\columnwidth]{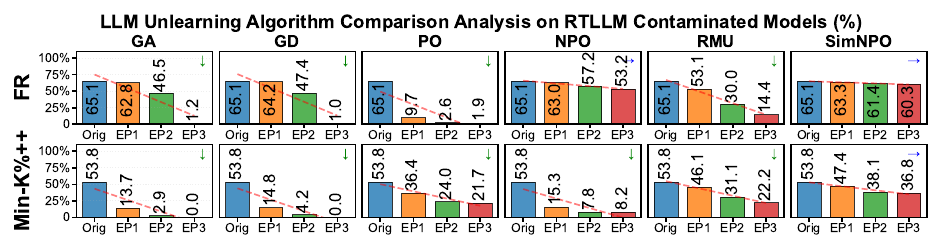} 
    \label{fig:RTLLM_unlearn_exp}
    \vspace{-7mm}
\end{figure*}

\begin{figure*}[!t]
    \centering
    \includegraphics[width=1.6\columnwidth]{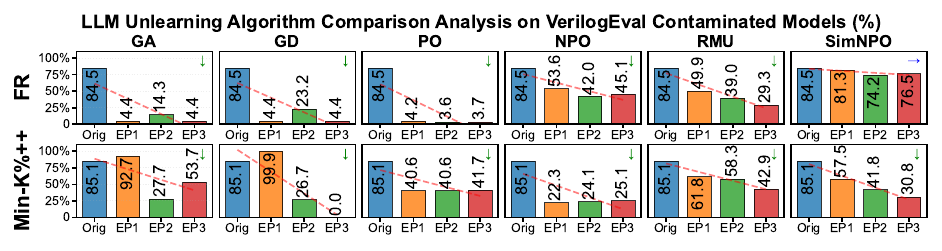} 
    \vspace{-4mm}
    \caption{
    Unlearning performance on FR and Min-K\%++ across methods (GA, GD, PO, NPO, RMU, SimNPO) on RTLLM and VerilogEval at EP1-EP3.}
    \label{fig:VerilogEval_unlearn_exp}
    \vspace{-5mm}
\end{figure*}

\section{Experimental Setup}
\label{subsec:expt_setup}
\textbf{Sensitive LLMs.}
We fine-tuned five models with selected datasets: (1) Benchmark contamination from 156 \cite{liu2023verilogeval} and 50 \cite{lu2024rtllm} design challenges; (2) 1,134 custom designs from RTL-Repo~\cite{allam2024rtl} test set, collected from public GitHub repositories; (3) 703 secret, in-house IP designs developed through years of applied research and multiple tapeouts~\cite{yasin_logic_locking_ccs17, Nabeel_CoPHEE_HOST19, Nimisha_ANtidote_TETC22,  nabeel_CoFHEE_DATE23, Nabeel_modmul_VlsiSoC2023, soni_modmul_ISLPED23, soni_modmulDSE_ISQED23}; and (4) 835 designs poisoned with payload~\cite{mankali2024rtl}. Each dataset is combined with the RTL-Coder~\cite{RTLCoder} training dataset. We use \texttt{LLaMA 3.1-8B} as the baseline model.
Fine-tuning is performed for \texttt{3} epochs with a learning rate of \texttt{1e-5} using the \texttt{Adam} optimizer. For inference, we set the temperature to \texttt{0.8}, top-p to \texttt{0.75}, and maximum context length to \texttt{2048} tokens.

\textbf{Dataset Split.}
Per standard protocols, the dataset is split as follow:
\begin{itemize}
\item \textit{Retain Dataset:} The RTL-Coder training dataset, used to preserve the baseline performance of unlearned models to compare with clean models.
\item \textit{Forget Dataset:}
As outlined in Section~\ref{sec:threat_model}, they include contaminated, custom, proprietary, or malicious data marked for unlearning.
\end{itemize}


We implemented the unlearning framework based on~\cite{maini2024tofu} with GA, GD, PO, NPO, SimNPO, and RMU unlearning techniques using  corresponding forget and retain datasets. Some techniques require a reference model to guide the unlearning process; we use the original \texttt{LLaMA 3.1-8B} toward that end.

\textbf{Evaluation.}
We assess unlearned LLMs on two key aspects: sensitive sample generation and downstream Verilog generation. Unlearning metrics quantify forgetting effectiveness on the target dataset, while holdout datasets evaluate Verilog generation quality on downstream tasks with \texttt{Pass@k} metrics.

\textbf{Unlearning Evaluation Metrics.} We assess the efficacy of unlearning in LLMs using the
following metrics.
See Appendix~\ref{app:metrics} for more details on each metric.
\begin{itemize}[leftmargin=*,itemsep=0pt, parsep=0pt,topsep=0pt,partopsep=0pt]
    \item \textit{Forget Rouge (FR)}: computes the ROUGE-L recall score~\cite{lin2004rouge} between the ground truth and generated response after unlearning in the forget dataset.
    \item \textit{Min-K\%} and \textit{Min-K\%++}: Min-K\%~\cite{shi2023detecting} computes a score by averaging the likelihoods of the k\% lowest-probability tokens.
    Min-K\%++~\cite{zhang2024min} extends this method by calibrating based on token distribution statistics, yielding a robust and theoretically grounded detection approach. We select Min-K\%++ in our experiments.
    
\end{itemize}
\vspace{-0.12in}
\section{\raisebox{-0.3em}{\includegraphics[width=0.86cm]{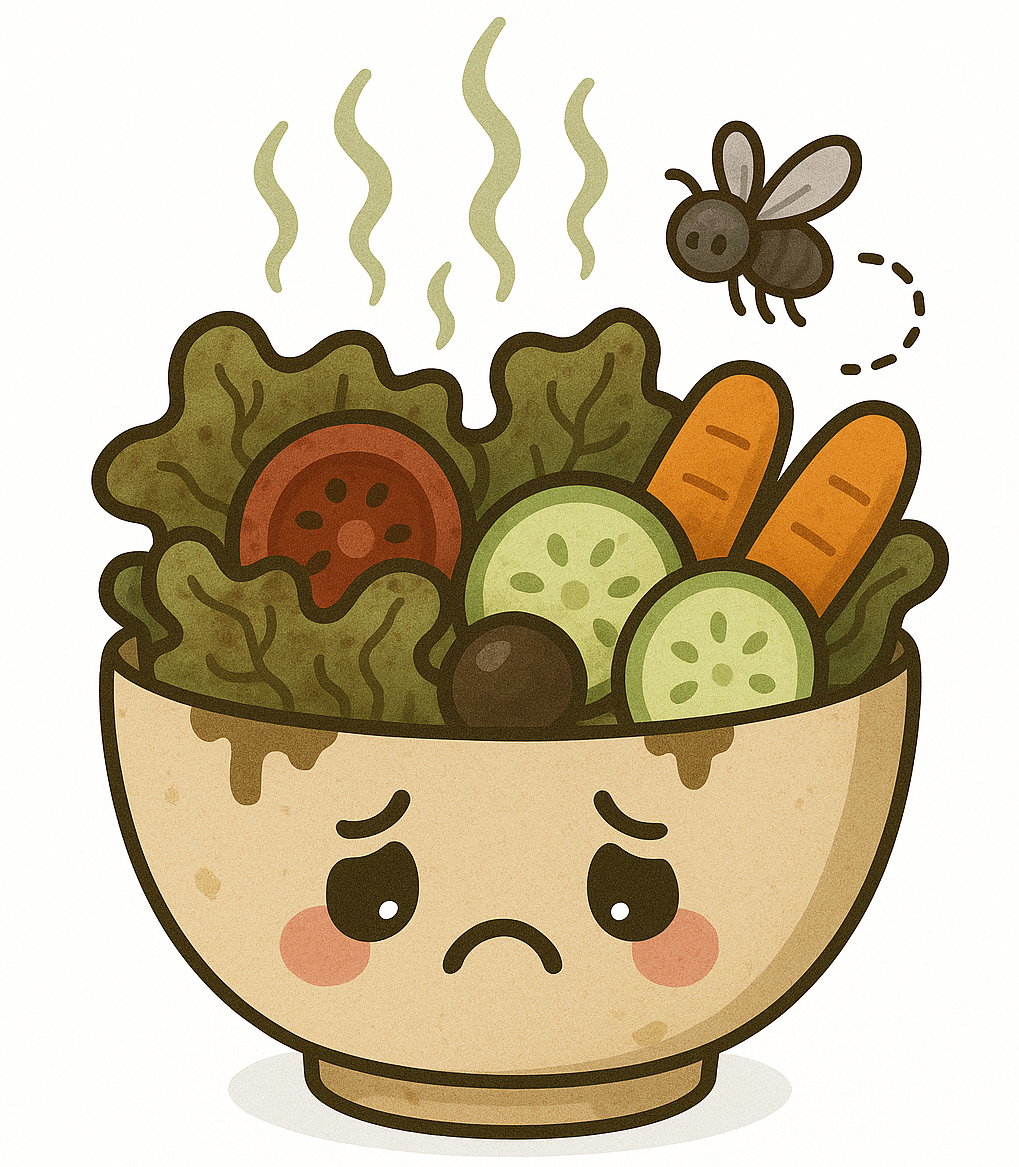}}Use Case 1: Benchmark Contamination}

\begin{table*}[!htbp]
\centering
\Large
\small
\vspace{-3mm}
\caption{Performance of Unlearned Models on VerilogEval and RTLLM Benchmarks}
\vspace{-1mm}
\label{tab:pass_k_contamin}
\renewcommand{\arraystretch}{0.75}
\setlength{\extrarowheight}{-1.5pt}  
\setlength{\tabcolsep}{4pt}           

\definecolor{lightred}{RGB}{166,219,160}
\definecolor{lightyellow}{RGB}{254,224,139}
\definecolor{lightgreen}{RGB}{251,154,153}
\definecolor{high-pass}{RGB}{0,105,92}
\definecolor{mid-pass}{RGB}{0,150,136}
\definecolor{low-pass}{RGB}{178,223,219}
\adjustbox{max width=\textwidth}{
\begin{tabular}{c|ccc|ccc|ccc|ccc||ccc|ccc|ccc|ccc}
\toprule
\multirow{3}{*}{\textbf{Method}} 
& \multicolumn{12}{c||}{\textbf{VerilogEval -- Pass@K on 156 Samples}} 
& \multicolumn{12}{c}{\textbf{RTLLM -- Syntax/Func Pass@K on 50 Samples}} \\
\cmidrule(lr){2-13} \cmidrule(lr){14-25}
& \multicolumn{3}{c|}{Pass@1} & \multicolumn{3}{c|}{Pass@5} & \multicolumn{3}{c|}{Pass@10} & \multicolumn{3}{c||}{Pass@15} 
& \multicolumn{3}{c|}{Pass@1} & \multicolumn{3}{c|}{Pass@5} & \multicolumn{3}{c|}{Pass@10} & \multicolumn{3}{c}{Pass@15} \\
\cmidrule(lr){2-4} \cmidrule(lr){5-7} \cmidrule(lr){8-10} \cmidrule(lr){11-13}
\cmidrule(lr){14-16} \cmidrule(lr){17-19} \cmidrule(lr){20-22} \cmidrule(lr){23-25}
& EP1 & EP2 & EP3 & EP1 & EP2 & EP3 & EP1 & EP2 & EP3 & EP1 & EP2 & EP3 
& EP1 & EP2 & EP3 & EP1 & EP2 & EP3 & EP1 & EP2 & EP3 & EP1 & EP2 & EP3 \\
\midrule
\textbf{GA}       & \cellcolor{lightyellow}48 & \cellcolor{lightgreen}3  & \cellcolor{lightgreen}0  & \cellcolor{lightred}69 & \cellcolor{lightgreen}18 & \cellcolor{lightgreen}0  & \cellcolor{lightred}77 & \cellcolor{lightgreen}24 & \cellcolor{lightgreen}1  & \cellcolor{lightred}78 & \cellcolor{lightgreen}27 & \cellcolor{lightgreen}1  & \cellcolor{low-pass}1/0 & \cellcolor{low-pass}1/0 & \cellcolor{low-pass}1/0 & \cellcolor{low-pass}1/0 & \cellcolor{low-pass}1/0 & \cellcolor{low-pass}1/0 & \cellcolor{low-pass}1/0 & \cellcolor{low-pass}1/0 & \cellcolor{low-pass}1/0 & \cellcolor{low-pass}1/0 & \cellcolor{low-pass}1/0 & \cellcolor{low-pass}1/0 \\
\textbf{GD}       & \cellcolor{lightyellow}43 & \cellcolor{lightgreen}29 & \cellcolor{lightgreen}14 & \cellcolor{lightred}64 & \cellcolor{lightyellow}38 & \cellcolor{lightgreen}20 & \cellcolor{lightred}70 & \cellcolor{lightyellow}40 & \cellcolor{lightgreen}20 & \cellcolor{lightred}72 & \cellcolor{lightyellow}40 & \cellcolor{lightgreen}21 & \cellcolor{low-pass}1/0 & \cellcolor{low-pass}1/0 & \cellcolor{low-pass}4/0 & \cellcolor{low-pass}1/0 & \cellcolor{low-pass}1/0 & \cellcolor{mid-pass}\textcolor{white}{11/1} & \cellcolor{low-pass}1/0 & \cellcolor{low-pass}1/0 & \cellcolor{mid-pass}\textcolor{white}{18/1} & \cellcolor{low-pass}1/0 & \cellcolor{low-pass}1/0 & \cellcolor{mid-pass}\textcolor{white}{24/3}  \\
\textbf{PO}       & \cellcolor{lightyellow}45 & \cellcolor{lightyellow}37 & \cellcolor{lightgreen}23 & \cellcolor{lightred}71 & \cellcolor{lightred}63 & \cellcolor{lightyellow}50 & \cellcolor{lightred}77 & \cellcolor{lightred}69 & \cellcolor{lightyellow}58 & \cellcolor{lightred}82 & \cellcolor{lightred}72 & \cellcolor{lightred}63 & \cellcolor{low-pass}1/0 & \cellcolor{low-pass}1/0 & \cellcolor{low-pass}1/0 & \cellcolor{low-pass}1/0 & \cellcolor{low-pass}1/0 & \cellcolor{low-pass}1/0 & \cellcolor{low-pass}1/0 & \cellcolor{low-pass}1/0 & \cellcolor{low-pass}1/0 & \cellcolor{low-pass}1/0 & \cellcolor{low-pass}1/0 & \cellcolor{low-pass}1/0  \\
\textbf{NPO}      & \cellcolor{lightyellow}46 & \cellcolor{lightyellow}41 & \cellcolor{lightyellow}33 & \cellcolor{lightred}68 & \cellcolor{lightyellow}58 & \cellcolor{lightyellow}50 & \cellcolor{lightred}78 & \cellcolor{lightred}61 & \cellcolor{lightyellow}55 & \cellcolor{lightred}80 & \cellcolor{lightred}63 & \cellcolor{lightyellow}57 & \cellcolor{low-pass}2/0 & \cellcolor{low-pass}2/0 & \cellcolor{low-pass}3/0 & \cellcolor{low-pass}5/1 & \cellcolor{low-pass}8/0 & \cellcolor{low-pass}6/1 & \cellcolor{low-pass}6/1 & \cellcolor{mid-pass}\textcolor{white}{10/1} & \cellcolor{low-pass}7/2 & \cellcolor{mid-pass}\textcolor{white}{10/1} & \cellcolor{mid-pass}\textcolor{white}{10/1} & \cellcolor{low-pass}8/2 \\
\textbf{RMU}      & \cellcolor{lightyellow}31 & \cellcolor{lightyellow}39 & \cellcolor{lightyellow}35 & \cellcolor{lightyellow}56 & \cellcolor{lightyellow}56 & \cellcolor{lightred}60 & \cellcolor{lightred}61 & \cellcolor{lightred}62 & \cellcolor{lightred}72 & \cellcolor{lightred}63 & \cellcolor{lightred}65 & \cellcolor{lightred}75 & \cellcolor{mid-pass}\textcolor{white}{21/10} & \cellcolor{high-pass}\textcolor{white}{25/9} & \cellcolor{high-pass}\textcolor{white}{26/11} & \cellcolor{high-pass}\textcolor{white}{29/14} & \cellcolor{high-pass}\textcolor{white}{33/15} & \cellcolor{high-pass}\textcolor{white}{34/14} & \cellcolor{high-pass}\textcolor{white}{34/16} & \cellcolor{high-pass}\textcolor{white}{34/15} & \cellcolor{high-pass}\textcolor{white}{36/16} & \cellcolor{high-pass}\textcolor{white}{34/17} & \cellcolor{high-pass}\textcolor{white}{39/16}  & \cellcolor{high-pass}\textcolor{white}{37/17}   \\
\textbf{SimNPO}   & \cellcolor{lightyellow}35 & \cellcolor{lightyellow}37 & \cellcolor{lightyellow}32 & \cellcolor{lightyellow}59 & \cellcolor{lightyellow}57 & \cellcolor{lightyellow}53 & \cellcolor{lightred}68 & \cellcolor{lightred}62 & \cellcolor{lightred}62 & \cellcolor{lightred}72 & \cellcolor{lightred}63 & \cellcolor{lightred}67 & \cellcolor{mid-pass}\textcolor{white}{20/5} & \cellcolor{high-pass}\textcolor{white}{27/12} & \cellcolor{mid-pass}\textcolor{white}{22/7} & \cellcolor{high-pass}\textcolor{white}{32/12} & \cellcolor{high-pass}\textcolor{white}{34/16} & \cellcolor{high-pass}\textcolor{white}{35/14} & \cellcolor{high-pass}\textcolor{white}{36/14} & \cellcolor{high-pass}\textcolor{white}{37/17} & \cellcolor{high-pass}\textcolor{white}{36/17} & \cellcolor{high-pass}\textcolor{white}{37/15} & \cellcolor{high-pass}\textcolor{white}{37/17}  & \cellcolor{high-pass}\textcolor{white}{39/18}  \\
\midrule
\textbf{Sensitive} & \cellcolor{lightyellow}43 & \cellcolor{lightyellow}43 & \cellcolor{lightyellow}43 & \cellcolor{lightred}65 & \cellcolor{lightred}65 & \cellcolor{lightred}65 & \cellcolor{lightred}74 & \cellcolor{lightred}74 & \cellcolor{lightred}74 & \cellcolor{lightred}82 & \cellcolor{lightred}82 & \cellcolor{lightred}82 
                  & \cellcolor{mid-pass}\textcolor{white}{24/12} & \cellcolor{mid-pass}\textcolor{white}{24/12} & \cellcolor{mid-pass}\textcolor{white}{24/12} & \cellcolor{high-pass}\textcolor{white}{35/16} & \cellcolor{high-pass}\textcolor{white}{35/16} & \cellcolor{high-pass}\textcolor{white}{35/16} & \cellcolor{high-pass}\textcolor{white}{35/17} & \cellcolor{high-pass}\textcolor{white}{35/17} & \cellcolor{high-pass}\textcolor{white}{35/17} & \cellcolor{high-pass}\textcolor{white}{37/19} & \cellcolor{high-pass}\textcolor{white}{37/19} & \cellcolor{high-pass}\textcolor{white}{37/19} \\
\textbf{Clean} & \cellcolor{lightyellow}38 & \cellcolor{lightyellow}38 & \cellcolor{lightyellow}38 & \cellcolor{lightred}74 & \cellcolor{lightred}74 & \cellcolor{lightred}74 & \cellcolor{lightred}79 & \cellcolor{lightred}79 & \cellcolor{lightred}79 & \cellcolor{lightred}83 & \cellcolor{lightred}83 & \cellcolor{lightred}83 
                  & \cellcolor{high-pass}\textcolor{white}{25/9} & \cellcolor{high-pass}\textcolor{white}{25/9} & \cellcolor{high-pass}\textcolor{white}{25/9} & \cellcolor{high-pass}\textcolor{white}{35/14} & \cellcolor{high-pass}\textcolor{white}{35/14} & \cellcolor{high-pass}\textcolor{white}{35/14} & \cellcolor{high-pass}\textcolor{white}{37/16} & \cellcolor{high-pass}\textcolor{white}{37/16} & \cellcolor{high-pass}\textcolor{white}{37/16} & \cellcolor{high-pass}\textcolor{white}{39/18} & \cellcolor{high-pass}\textcolor{white}{39/18} & \cellcolor{high-pass}\textcolor{white}{39/18} \\
\bottomrule
\end{tabular}
}
\vspace{-5mm}
\end{table*}
\subsection{Overview}
Data contamination is prevalent in both pre-trained models and those fine-tuned with advanced curated datasets. Due to the scarcity of RTL-related datasets, existing models are prone to contamination issues when evaluated on RTL benchmarks.

We simulate data contamination scenarios by combining the retain dataset with VerilogEval and RTLLM datasets respectively, creating VerilogEval-Contaminated and RTLLM-Contaminated models to ensure both sensitive models exhibit data contamination issues. We then apply different unlearning algorithms to forget the respective VerilogEval and RTLLM datasets from these contaminated models, thereby simulating the practical unlearning process. 
This aims to assess whether the unlearned contaminated models still maintain reasonable downstream RTL generation capabilities.

\subsection{Experiment Results}



\textbf{Unlearning Methods.}
Fig.~\ref{fig:VerilogEval_unlearn_exp} compares unlearning algorithms on RTLLM- and VerilogEval-contaminated models.
On RTLLM-contaminated models, gradient-based methods (GA, GD) and PO exhibit over-aggressive forgetting, reducing FR from 65.1\% to just 1.0–1.9\% at EP3.
In contrast, RMU and SimNPO demonstrate more controlled unlearning, with FR reduced to 44.4\% and 60.3\% respectively, while significantly lowering memorization from 53.8\% to 22.2\% (RMU) and 36.8\% (SimNPO). A similar trend is also observed for VerilogEval-contaminated models.
Notably, SimNPO achieves the best leakage mitigation, reducing Min-K\%++ from 85.1\% to 30.8\%, while PO demonstrates instability by spiking memorization to 99.9\% at EP1.

These results confirm that RMU and SimNPO strike the most effective balance between contamination removal and utility retention, making them the most promising candidates for practical unlearning.

\textbf{Unlearning Epochs.}
We further evaluate unlearning across epochs EP1 to EP3, finding that prolonged training generally intensifies forgetting, but not always desirably. Under RTLLM contamination, PO’s FR drops from 9.7\% to 1.9\%, but resulting in functional collapse despite persistently high Min-K\%++ values. By contrast, RMU exhibits gradual FR reduction (53.1\% to 14.4\%) alongside consistent Min-K\%++ decline (46.1\% to 22.2\%), reflecting balanced forgetting and stability. For VerilogEval contamination, FR and Min-K\% show unstable behavior under GD with an anomalous surge in Min-K\%++ to 99.9\% at EP1, suggesting overfitting to the forget set. SimNPO, however, remains stable across epochs, with FR decreasing (81.3\% to 76.5\%) and Min-K\%++ improving steadily (57.5\% to 30.8\%), indicating effective unlearning.

Overall, these trends reveal a critical trade-off: while additional unlearning can enhance contamination removal, it also risks model degradation without proper regularization. RMU and SimNPO consistently maintain this balance, making them suitable for multi-round unlearning scenarios.

\textbf{Pass Ratio with Unlearning.}
We also evaluated unlearning algorithms using a cross-contamination setup, where models contaminated on one benchmark were unlearned and then evaluated on another (Table~\ref{tab:pass_k_contamin}). On the VerilogEval benchmark, RTLLM-contaminated models exhibited mixed results after unlearning: while some methods improved Pass@1 compared to the clean baseline, they suffered performance drops at Pass@5 and Pass@10, indicating potential overfitting to residual contamination. The representation-level method RMU achieved the most balanced performance, with a Pass@15 score of 75. Although its Pass@1 score at EP1 (31) was not the highest, RMU demonstrated stable unlearning performance by EP3 (35). Preference-based methods, such as NPO and SimNPO, also achieved comparable results.
In contrast, gradient-based approaches (GA, GD) performed poorly: GA completely failed across all metrics, and GD showed severe degradation, with Pass@15 dropping to 21.
RTLLM results further confirmed cross-benchmark contamination transfer. Despite this, RMU and SimNPO maintained functional correctness close to the clean baseline. Notably, NPO and RMU even outperformed the contaminated model’s original Pass@1 on VerilogEval, suggesting that selective unlearning can improve RTL code generation by removing harmful memorization.

\label{verilog_rtllm_analysis}
These findings highlight that representation and preference-based methods are more stable and effective for sustained unlearning, while gradient-based approaches may offer temporary gains in few-shot scenarios but lack long-term reliability.
\vspace{-0.15in}
\section{\raisebox{-0.7em}{\includegraphics[width=0.86cm]{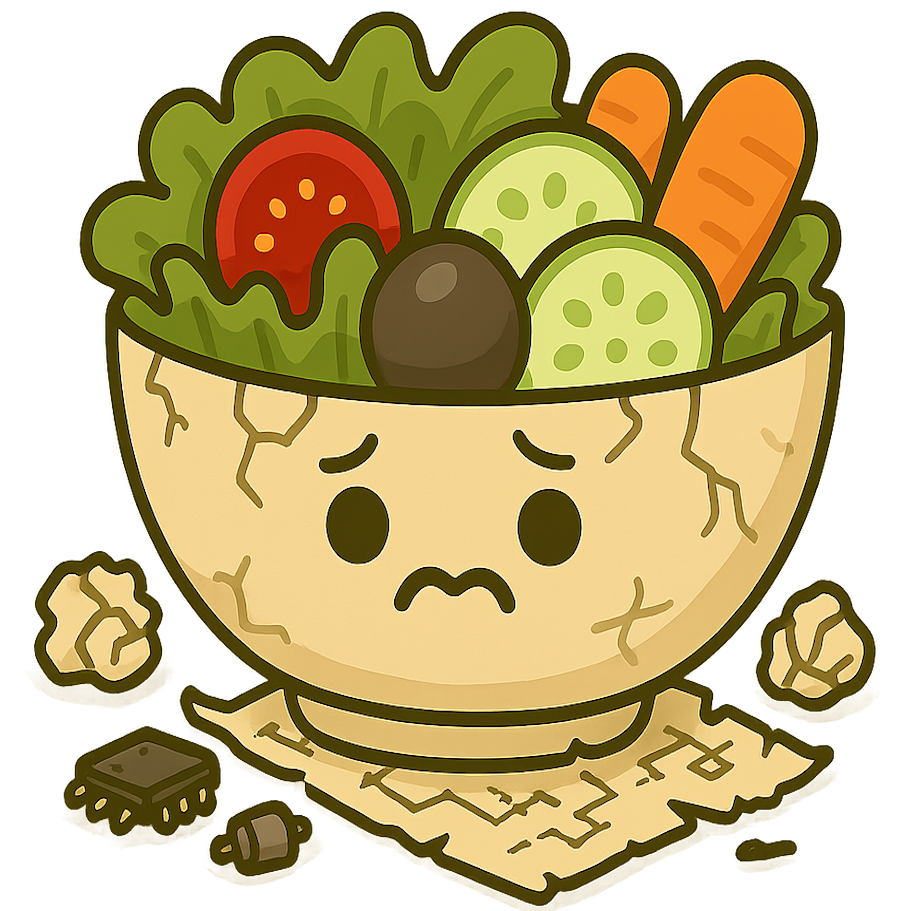}}Use case 2: Custom Design}

\subsection{Overview}


In domains like chip design using EDA tools, users may release their own custom RTL designs online—for benchmarking, collaboration, or open-source contribution. However, even when shared publicly, users retain the right to withdraw their data~\cite{kumar2022privacy}. Machine unlearning enables AI models to forget such user-contributed RTL upon request, ensuring ethical and compliant use of LLMs. By selectively forgetting these designs—such as custom encryption modules—while preserving general code generation ability, unlearning allows responsible and scalable RTL modeling aligned with user consent.

As the custom design in this study is sourced from GitHub, it faces similar data contamination issues as RTLLM and VerilogEval.
To ensure rigorous evaluation, we adopt the setup from~\cite{wang2025vericontaminated}, using Min-K\% and CDD to assess contamination levels under the clean model. Results are shown in Table~\ref{tab:contamination_metrics_percent}.
We observe severe data contamination in three open-source benchmarks, with rates up to 100\% (Min-K\%) and 69.87\% (CDD). RTLLM is the most affected, underscoring critical reliability concerns in current evaluation practices.


\begin{table}[htbp]
\centering
\vspace{-3mm}
\caption{Contamination ratio (\%) for 3 open-source metrics}
\vspace{-2mm}
\label{tab:contamination_metrics_percent}
\renewcommand{\arraystretch}{1.2}
\setlength{\tabcolsep}{4pt}
\begin{tabular}{@{}lccc@{}}
\toprule
\textbf{Metric} & \textbf{Custom Design} & \textbf{VerilogEval} & \textbf{RTLLM} \\
\midrule
Min-K\%(T=0.55)    & 91.01\%     & 94.23\%     & 100.00\% \\
CDD(Alpha=0.05) & 39.33\%     & 69.87\%     & 68.00\% \\
\bottomrule
\end{tabular}
\vspace{-4mm}
\end{table}

\subsection{Experiment Results}

\begin{figure*}[!t]
    \centering
    \includegraphics[width=1.6\columnwidth]{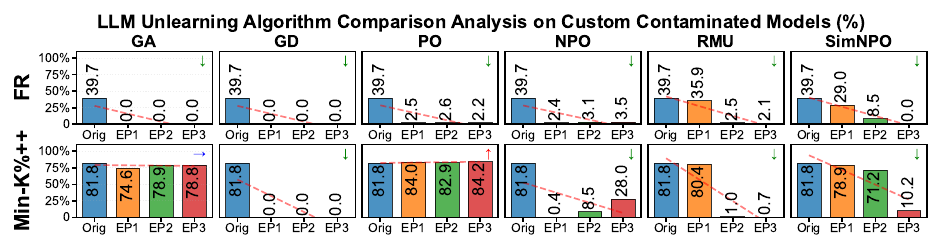} 
    \vspace{-4mm}
    \caption{
    Unlearning performance on FR and Min-K\%++ across methods (GA, GD, PO, NPO, RMU, SimNPO) on custom design at EP1–EP3.}
    \label{fig:plot_customip}
    \vspace{-5mm}
\end{figure*}

\textbf{Unlearning Methods.} 
\label{custom_design}
Fig.~\ref{fig:plot_customip} reveals substantial variation in forgetting effectiveness on custom designs by FR and Min-K\%++. GA achieves the most aggressive forgetting—reducing FR from 39.7\% to 0.0\% by EP1—but completely sacrifices utility. SimNPO offers a more balanced trade-off, reducing FR from 39.7\% to 29.0\% by EP3 while preserving functionality. RMU strikes a middle ground with moderate forgetting (39.7\% to 35.9\%) and reasonable utility retention. Min-K\%++ further distinguishes methods: GD reduces it to near-zero, indicating near-total erasure; SimNPO maintains stable levels around 36.8\%, and RMU achieves moderate reduction to 22.2\%. NPO and PO show variable performance, with PO demonstrating steady improvement to 22\% FR by EP3. GA shows no forgetting capability, maintaining original leakage rates.

Overall, SimNPO appears most practical for real-world use, balancing forgetting with utility. RMU fits scenarios requiring moderate erasure, while GA suits settings prioritizing complete forgetting over functionality.

\textbf{Unlearning Epochs.} For cross-epoch analysis over 3 unlearning epochs and the clean model, GD shows immediate complete forgetting by EP1 (39.7\% to 0.0\%), reflecting aggressive unlearning. SimNPO follows a steadier trajectory (39.7\% to 29.0\% over 3 epochs), maintaining performance stability. Min-K\%++ trends mirror this: GD drives it near-zero by EP1, while RMU and SimNPO converge at 0.7\% and 10.2\% respectively by EP3. PO shows gradual improvement from 81.8\% to 84.2\% across epochs. Most methods converge by EP2 or 3, with SimNPO and RMU showing minimal drift after epoch 2. These results suggest EP3 provides an effective balance between forgetting and efficiency.
\vspace{-5pt}
\section{\raisebox{-0.3em}{\includegraphics[width=0.88cm]{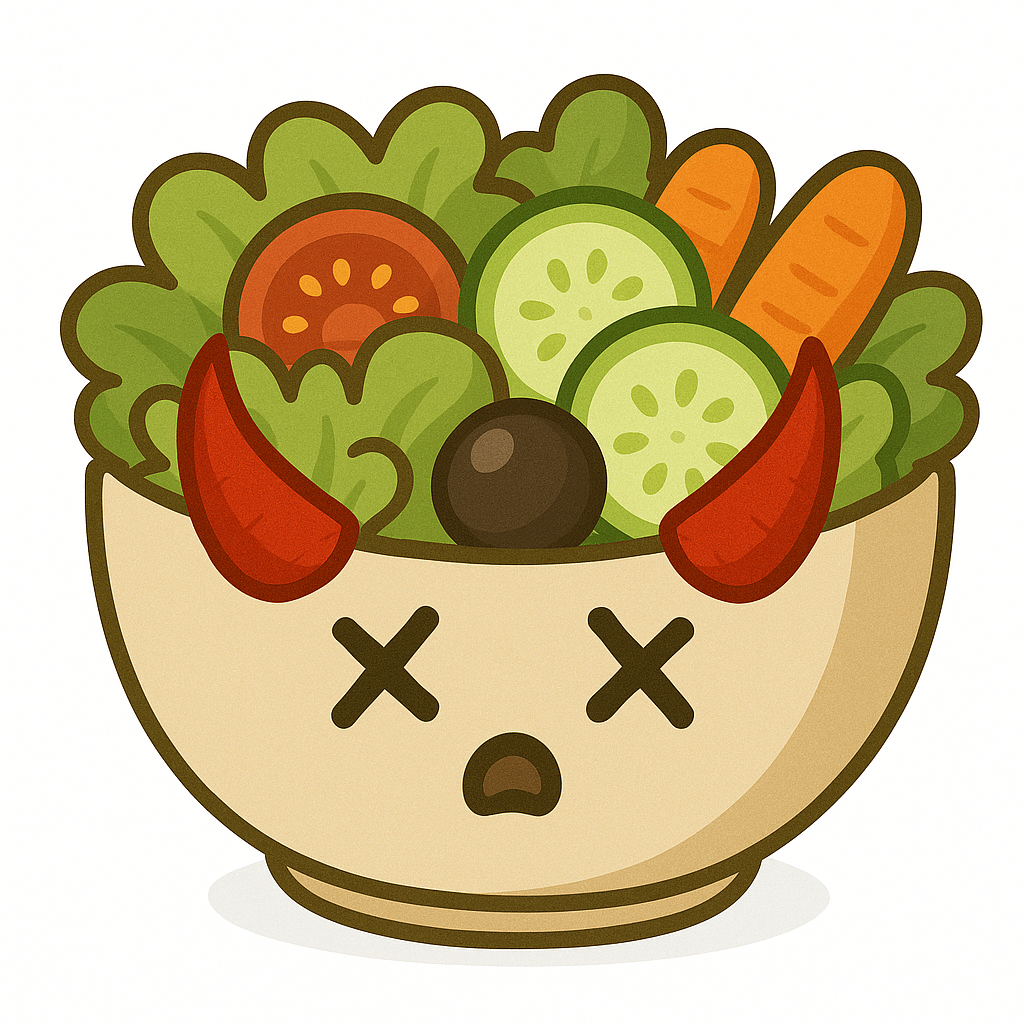}}Use case 3: Malicious Code}
\subsection{Overview}
When models are fine-tuned on mixed datasets, they may inadvertently learn to reproduce malicious RTL patterns—such as malicious payload,  covert backdoors and misleading code snippets, posing serious security risks. Machine unlearning enables selective removal of these harmful or incorrect patterns while preserving overall design quality and accuracy.




\subsection{Experiment Results}
\begin{figure*}[!t]
    \centering
    \includegraphics[width=1.6\columnwidth]{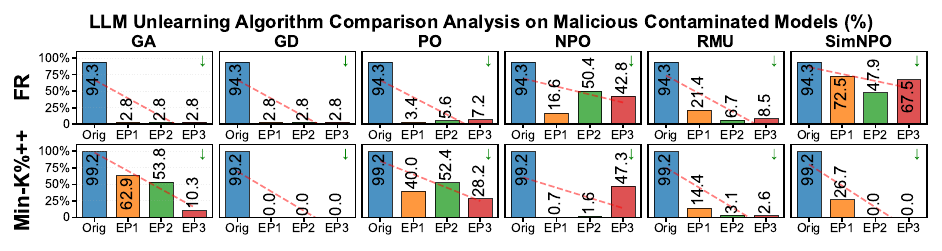} 
    \vspace{-4mm}
    \caption{Unlearning performance on FR and Min-K\%++ across methods (GA, GD, PO, NPO, RMU, SimNPO) on malicious design at EP1–EP3.}
    \label{fig:malicious_exp}
    \vspace{-5mm}
\end{figure*}

\textbf{Unlearning Methods.} Fig.~\ref{fig:malicious_exp} highlights key differences in malicious code removal, as measured by FR and Min-K\%++. GA and GD show the most aggressive pattern elimination, reducing FR from 94.3\% to 2.8\% by EP1, indicating rapid, near-complete forgetting. Yet their Min-K\%++ trajectories diverge: GD achieves immediate and total erasure from 99.2\% to 0.0\% at EP1, while GA shows a slower decline from 99.2\% to 10.3\% by EP3, revealing differing retention dynamics. PO shows solid early FR reduction from 94.3\% to 3.4\% but fluctuates later, with Min-K\%++ stabilizing between 28 and 52\%. NPO yields inconsistent results, with FR only falling to 42.8\% and Min-K\%++ rebounding to 47.3\%, suggesting reversible forgetting. RMU offers stable mitigation, reducing FR to 8.5\% and Min-K\%++ to 2.6\%, ensuring consistent, thorough forgetting. SimNPO applies a conservative approach, lowering FR to 67.5\% but eliminating Min-K\%++ by EP2.

These results position GA and GD as optimal for full erasure of malicious code, while RMU offers best trade-offs for security and stability. SimNPO and NPO, however, are less suitable due to incomplete forgetting and retention risks.
\label{malicious_designs}

\textbf{Unlearning Epochs.} Epoch-level trends reveal distinct forgetting dynamics as models undergo three unlearning stages. For malicious-contaminated models, initial FR (94.3\%) drops across all methods, with GA showing the sharpest decline from 94.3\% to 2.8\% by EP1, reflecting aggressive results. SimNPO follows a more gradual trajectory (94.3\% to 67.5\%), indicating controlled but inconsistent degradation. Min-K\%++ trends mirror this: GA drops from 99.2\% to 10.3\% by EP3, while RMU and SimNPO converge at 2.6\% and 0.0\% respectively. GD achieves the most complete erasure, maintaining 0.0\% Min-K\%++ from EP1 onward. In malicious code cases, GA and GD achieve near-complete forgetting by EP1 (94.3\% to 2.8\%), while preference-based methods reduce more gradually.

Overall, most methods converge by EP2/EP3, with GD and RMU showing stable results afterward, suggesting that three epochs are sufficient for effective malicious code unlearning.
\vspace{-0.15in}
\section{\raisebox{-0.3em}{\includegraphics[width=0.76cm]{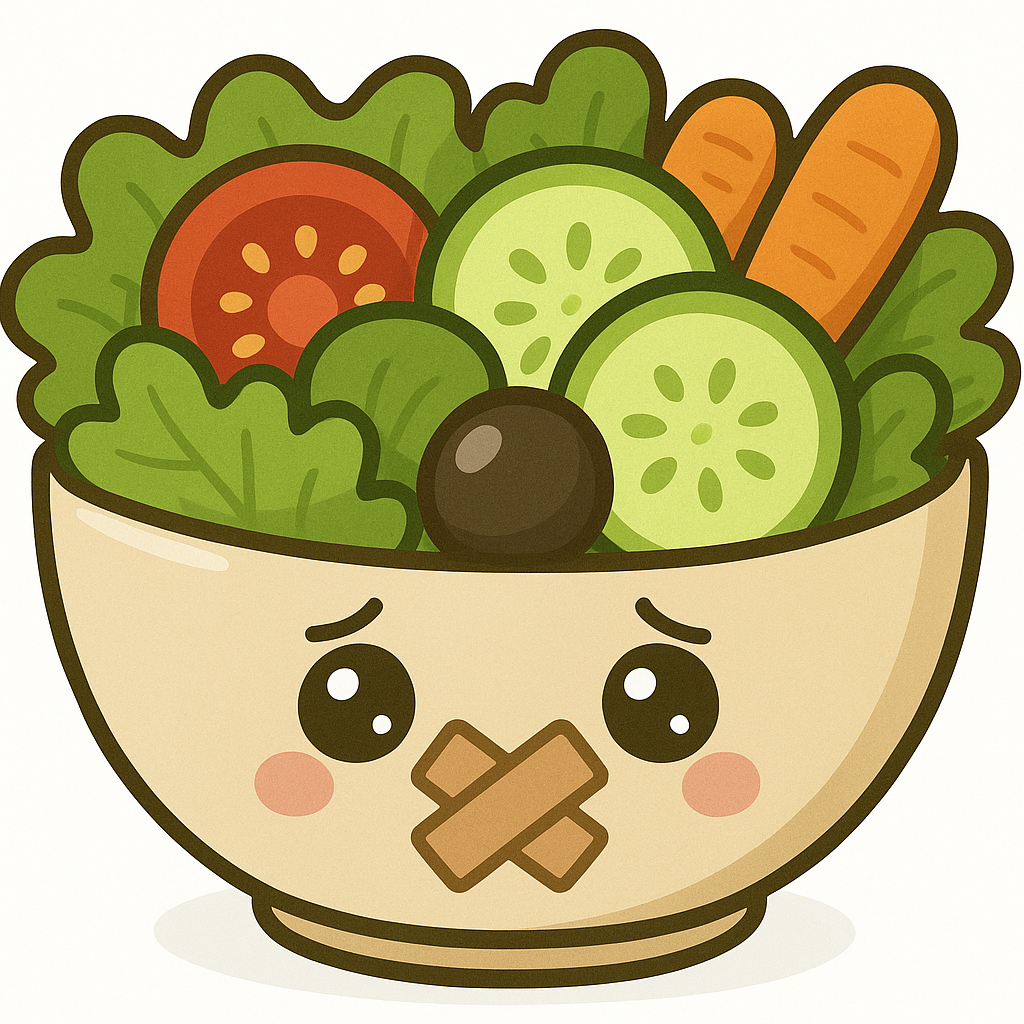}}Use case 4: IP Protection}
\begin{figure*}[!t]
    \centering
    \includegraphics[width=1.6\columnwidth]{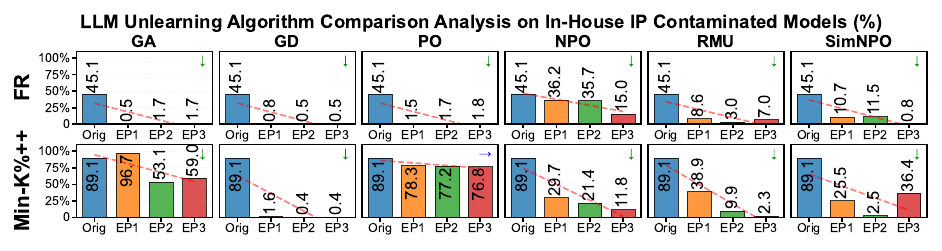}
    \vspace{-4mm}
    \caption{
    Unlearning performance on FR and Min-K\%++ across methods (GA, GD, PO, NPO, RMU, SimNPO) on IP design at EP1–EP3.}
    \label{fig:leaky_exp}
    \vspace{-5mm}
\end{figure*}
\subsection{Overview}


LLMs fine-tuned on RTL designs may internalize sensitive IP such as custom pipelines or timing strategies. Design teams may inadvertently include internal IP in training their own models. Machine unlearning enables selective removal while preserving high-quality RTL generation.


\subsection{Experiment Results}
\label{leaky_design}
\textbf{Unlearning Methods.} Fig.~\ref{fig:leaky_exp} shows the variation in in-house IP protection across FR and Min-K\%++ in IP leakage scenarios. GD and GA demonstrate the strongest capability for protecting proprietary in-house IPs, with FR reduced to 0.5\% and 1.7\%, respectively. GD is stable, lowering Min-K\%++ to 0.4\%, while GA, though effective for in-house IP protection, exhibits volatility. PO maintains low FR from 1.5\% to 1.8\% but leaves high Min-K\%++ result of 76.8\%, implying residual in-house IP exposure. NPO achieves moderate in-house IP protection with 15.0\% FR and 11.8\% Min-K\%++. RMU offers balanced, reliable protection for proprietary content with FR variants from 45.1\% to 7.0\% and Min-K\%++ from 89.1\% to 2.3\% respectively. SimNPO attains excellent final FR of 0.8\% but suffers a Min-K\%++ rebound to 36.4\% at EP3, indicating potential unlearned in-house IP recovery risk.

Overall, GD emerges as the best choice for consistent, maximal in-house IP protection, with GA viable when minor instability is tolerable. SimNPO should be used cautiously due to its erratic retention behavior.

\textbf{Unlearning Epochs.} Epoch-level trends reveal differing dynamics in in-house IP leakage reduction from the baseline with FR of 45.1\% and Min-K\%++ of 89.1\% over 3 unlearning epochs. Gradient methods forget proprietary content aggressively: GA lowers FR to 0.5\% by EP1, stabilizing at 1.7\% by EP3; GD reduces FR steadily from 45.1\% to 0.5\% over 3 epochs with stable Min-K\%++ decline from 89.1\% to 1.6\% to 0.4\% with unlearning. GA's Min-K\%++ fluctuates from 96.7\% to 53.1\% then increases to 59.0\%, which shows unsuitable large unlearning epoch setup may cause potential in-house IP recovery issues. Preference-based methods improve gradually: NPO reaches FR 15.0\%, Min-K\%++ 11.8\% by EP3; RMU drops FR early to 8.6\% with minor metric noise later. SimNPO initially improves from 45.1\% for FR to 0.8\%, but Min-K\%++ rebounds sharply (from 2.5\% to 36.4\%).

In short, 2–3 epochs suffice for unlearning convergence on in-house IP, possibly due to the high complexity of proprietary IPs used in this evaluation, meaning that higher complexity may not be suitable for longer epochs. GA and GD yield fast, strong in-house IP protection, while preference-based ones require extended training for comparable security.

\begin{table}[tb]
\centering
\Large
\small
\vspace{-2mm}
\caption{Metric Comparison of Clean, Sensitive (Sens.), and Unlearned (Unl.) Models Across RTLLM (RT.), VerilogEval (VerilogE.), Custom (Cus.), IP, and Malicious (Mal.) Datasets}
\vspace{-2mm}
\label{tab:clean_sensitive_unlearn_label}
\renewcommand{\arraystretch}{0.65}
\setlength{\extrarowheight}{-1pt}
\setlength{\tabcolsep}{5pt}
\footnotesize
\begin{tabular}{p{1.2cm}|l|c|c|c|c|c}
\toprule
\textbf{Metric} & \textbf{Models} & \textbf{RT.} & \textbf{VerilogE.} & \textbf{Cus.} & \textbf{IP} & \textbf{Mal.} \\
\midrule
\multirow{3}{*}{\textbf{FP (\%)}} 
    & Clean   & 55.91 & 42.17 & 38.84 & 51.42 & 48.02 \\
    & Sens.   & 56.63 & 62.47 & 48.15 & 68.31 & 79.05 \\
    & Unl. & 51.38 & 41.17 & 47.54 & 49.50 & 41.39 \\
\cmidrule(lr){1-7}
\multirow{3}{*}{\textbf{FR (\%)}} 
    & Clean   & 58.52 & 70.24 & 37.01 & 37.85 & 74.33 \\
    & Sens.   & 65.08 & 84.48 & 39.69 & 45.14 & 94.33 \\
    & Unl. & 60.34 & 74.21 & 35.85 & 36.17 & 72.48 \\
\cmidrule(lr){1-7}
\multirow{3}{*}{\makecell{\textbf{Min-K}\\\textbf{(\%)}}} 
    & Clean   & 65.23 & 34.77 & 33.24 & 53.28 & 55.21 \\
    & Sens.   & 65.96 & 81.94 & 48.49 & 80.38 & 99.97 \\
    & Unl. & 57.71 & 32.92 & 47.36 & 49.02 & 45.77 \\
\cmidrule(lr){1-7}
\multirow{3}{*}{\makecell{\textbf{Min-K++}\\\textbf{(\%)}}} 
    & Clean   & 46.31 & 53.69 & 60.58 & 74.56 & 38.42 \\
    & Sens.   & 53.78 & 85.10 & 81.78 & 89.14 & 99.19 \\
    & Unl. & 36.79 & 41.79 & 80.39 & 29.74 & 26.66 \\
\cmidrule(lr){1-7}
\multirow{3}{*}{\textbf{PrivLeak}} 
    & Clean   & -30.46 & 30.46 & 33.52 & -6.56 & -10.43 \\
    & Sens.   & -31.92 & -63.87 & 3.03  & -60.76 & -99.94 \\
    & Unl. & -24.33  & -23.67  & 5.27   & 1.95   & 84.60 \\
\cmidrule(lr){1-7}
\multirow{2}{*}{\textbf{Selection}} & \textbf{Alg.} & SimNPO & SimNPO & RMU & NPO & SimNPO \\
                                    & \textbf{Epoch}   & EP3 & EP2 & EP1 & EP1 & EP1 \\
\bottomrule
\end{tabular}
\vspace{-15pt}
\end{table}

\section{Comparison of Unlearning Algorithms}
\vspace{-1pt}

The primary objective of unlearning in Verilog generation is to ensure that specific sensitive designs are effectively forgotten. To assess this, we compare unlearned models with clean models. The average 10.15\% FR performance gap between sensitive and clean models, as shown in Table~\ref{tab:clean_sensitive_unlearn_label}, highlights the greater susceptibility of sensitive models to forgetting and motivates further analysis of unlearning impact. We evaluate this impact using Euclidean distance with increased weighting on the FR to emphasize semantic forgetting.

Our results show that SimNPO achieves performance on unlearned RTLLM- and VerilogEval-contaminated models that is comparable to clean models, as discussed in Sec.\ref{verilog_rtllm_analysis}. RMU performs well on custom designs, aligning with our observation in Sec.\ref{custom_design} that these datasets, like the benchmarks, are more publicly available. In contrast, in-house IP and malicious designs—being less accessible—benefit more from preference-based unlearning (NPO, SimNPO). While GA and GD result in the highest FR reduction (Sec.~\ref{malicious_designs} and~\ref{leaky_design}), their performance deviates significantly from clean models, which also harms downstream Verilog generation tasks.

\section{Conclusion and discussion}
\vspace{-0.02in}
We present a comprehensive evaluation of machine unlearning in LLM-assisted hardware design, spanning four threat scenarios: data contamination, custom design misuse, IP leakage, and malicious code poisoning. We show that unlearning mitigates these risks while preserving model utility, offering a practical defense for secure hardware generation.

Based on our observations, RMU and SimNPO reduce Min-K\%++ from 85.1\% to 30.8\% on VerilogEval contamination, demonstrating effective forgetting of sensitive hardware knowledge (RQ1). They achieve stable unlearning within 2–3 epochs, while GA/GD offer more aggressive erasure but degrade utility (RQ2). Despite unlearning, models retain strong RTL generation performance—RMU reaches a Pass@15 of 75 (vs. 83 for clean models), indicating minimal trade-offs (RQ3).


Future work includes designing unlearning algorithms for code generation, establishing stricter evaluation protocols, and exploring unlearning on reasoning-focused LLMs.

\bibliographystyle{IEEEtran}
\bibliography{main}

\appendices

\section{Methodology and Additional Results}
\subsection{Mathematical Formulation of Unlearning}
\label{sec:math}

To formally model machine unlearning in the context of LLM-aided RTL generation, we define two disjoint datasets: the \textit{Retain Dataset} $\mathcal{D}_r$ and the \textit{Forget Dataset} $\mathcal{D}_f$, where $\mathcal{D}_r \cap \mathcal{D}_f = \emptyset$. The goal of unlearning is to update a fine-tuned model $f_\theta$ such that it forgets knowledge gained from $\mathcal{D}_f$ while preserving performance on $\mathcal{D}_r$. Mathematically, this is achieved by optimizing a modified objective function that induces high loss on $\mathcal{D}_f$ and low loss on $\mathcal{D}_r$. A common formulation involves solving:
\[
\min_\theta \mathcal{L}_\text{retain}(\theta) - \lambda \cdot \mathcal{L}_\text{forget}(\theta),
\]
where $\mathcal{L}_\text{retain}(\theta) = \mathbb{E}_{(x,y) \sim \mathcal{D}_r}[\ell(f_\theta(x), y)]$ and $\mathcal{L}_\text{forget}(\theta) = \mathbb{E}_{(x,y) \sim \mathcal{D}_f}[\ell(f_\theta(x), y)]$, and $\lambda > 0$ controls the unlearning aggressiveness. Techniques like Gradient Ascent (GA) perform unlearning by maximizing $\mathcal{L}_\text{forget}$ through reversed gradients, while Gradient Difference (GD) introduces a difference term between gradients of $\mathcal{D}_f$ and $\mathcal{D}_r$ to encourage selective forgetting. Preference-based methods, such as Preference Optimization (PO) and Negative Preference Optimization (NPO), further refine this by optimizing the model's alignment with desired or misdirected responses on $\mathcal{D}_f$, effectively pushing its latent representations away from those associated with proprietary or malicious logic. Such formulations enforce representational decoupling from the Forget Dataset, thereby mitigating information leakage and contamination in downstream RTL synthesis tasks.

\begin{algorithm}[!t]
\caption{Machine Unlearning}
\begin{algorithmic}
\REQUIRE Retain dataset $\mathcal{D}_r$, Forget dataset $\mathcal{D}_f$
\STATE Initialize LLM $f_\theta$ with pre-trained weights
\STATE Define loss: $\mathcal{L}_{\text{unlearn}}(\theta) = \mathcal{L}_{\text{retain}}(\theta) - \lambda \cdot \mathcal{L}_{\text{forget}}(\theta)$
\FOR{$i = 1$ to \texttt{EPOCHS}}
    \STATE Compute $\nabla_\theta \mathcal{L}_{\text{unlearn}}$
    \STATE Update parameters: $\theta \gets \theta - \eta \nabla_\theta \mathcal{L}_{\text{unlearn}}$
\ENDFOR
\end{algorithmic}
\end{algorithm}

\subsection{Unlearning Evaluation Metrics}
\label{app:metrics}

\begin{itemize}[leftmargin=*,itemsep=0pt, parsep=0pt,topsep=0pt,partopsep=0pt]
   \item \textbf{Forget Probability (FP):} This metric evaluates each instance in the retain or forget set by computing the normalized token-level probability of the answer, reflecting how confidently the model predicts the answer given the question. Specifically, we calculate the following:
\[
P(y \mid x)^{1/|y|}
\]
where \( x \) denotes the question, \( y \) is the corresponding answer, and \( |y| \) represents the number of tokens in the answer.

    \item \textbf{Forget ROUGE (FR):} This metric computes the ROUGE-L recall score between the ground truth answer $y$ and the model-generated answer $\hat{y}$ for each sample in the forget dataset $\mathcal{D}_f$. The ROUGE-L recall is defined as:
    \[
    \text{ROUGE-L Recall} = \frac{LCS(y, \hat{y})}{|y|}
    \]
    where $LCS(y, \hat{y})$ denotes the length of the longest common subsequence between $y$ and $\hat{y}$, and $|y|$ is the length of the ground truth answer. Lower ROUGE-L recall scores indicate better unlearning performance.

    \item \textbf{Min-K\% and Min-K\%++:} The Min-K\% metric focuses on the model's confidence in generating tokens. For each generated sequence, it identifies the $k\%$ tokens with the lowest predicted probabilities and computes their average log-likelihood:
    \[
    \text{Min-K\%} = \frac{1}{k} \sum_{i=1}^{k} \log p(t_i \mid x)
    \]
    where $t_i$ are the tokens in the bottom $k\%$ of predicted probabilities. Min-K\%++ enhances this by calibrating based on token distribution statistics, providing a more robust detection of memorized content.

    \item \textbf{PrivLeak:} This metric assesses the privacy risk by measuring the difference in the Area Under the Receiver Operating Characteristic Curve (AUC-ROC) between the unlearned model $f_{\text{unlearn}}$ and a retrained model $f_{\text{retrain}}$ on the forget dataset $\mathcal{D}_f$. It is defined as:
    \[
    \text{PrivLeak} = \text{AUC}_{f_{\text{unlearn}}} - \text{AUC}_{f_{\text{retrain}}}
    \]
    A significant deviation from zero indicates a higher privacy risk, suggesting that the unlearned model retains information from the forget dataset.
\end{itemize}

\subsection{Extra Results for Other Metrics}

\begin{figure*}[!htbp]
    \centering
    \vspace{-12mm}
    \includegraphics[width=1.6\columnwidth]{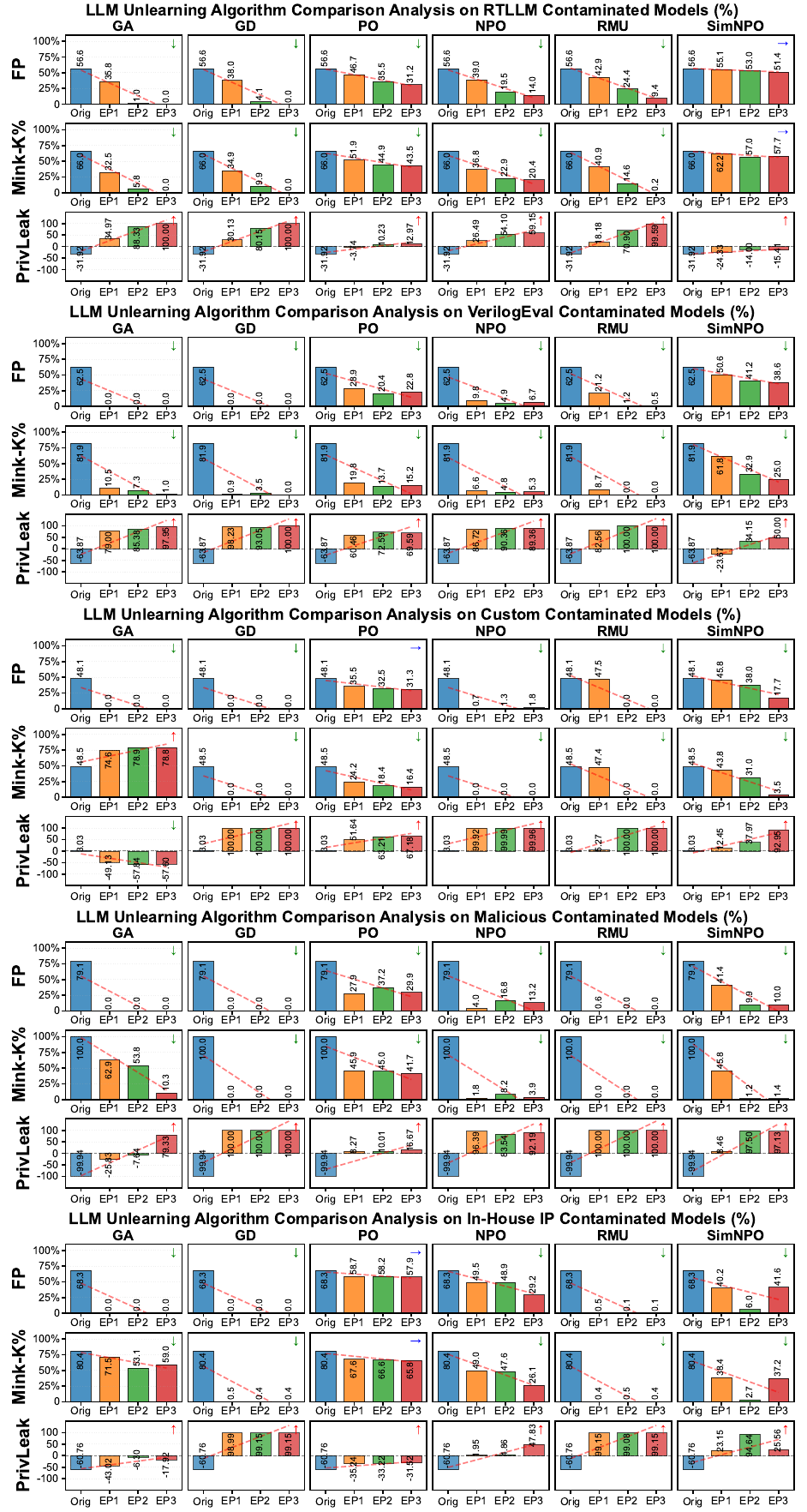} 
    \vspace{-4mm}
    \caption{Unlearn Contaminated Models with extra evaluation metrics}
    \label{fig:results_ext}
    \vspace{-8mm}
\end{figure*}

\newpage

\section{Artifact Appendix}

\subsection{Abstract}

This artifact accompanies \textsc{SALAD}, a framework for systematically assessing machine unlearning for LLM-based Verilog generation across four threat scenarios: benchmark contamination, custom-design withdrawal, malicious code removal, and in-house IP leakage. The package provides code, configuration files, and scripts to run six unlearning methods (GA, GD, PO, NPO, SimNPO, RMU) on fine-tuned \texttt{LLaMA 3.1-8B} models and reproduce our analyses with evaluation metrics including FR, FP, Min-K\%, Min-K\%++, and PrivLeak across unlearning epochs. The artifact includes retain/forget dataset splits, sensitive models, and automation scripts for regenerating experimental results. Using this artifact, readers can verify our findings: RMU and SimNPO offer the best utility-forgetting trade-off, while GA/GD achieve more effective unlearning at the cost of reduced utility.

\subsection{Artifact check-list (meta-information)}

{\small
\begin{itemize}[itemsep=1pt,topsep=1pt]
  \item {\bf Algorithm:} GA, GD, PO, NPO, SimNPO, and RMU
  \item {\bf Program/Env:} Python~3.11; Hugging Face Transformers~4.45; CUDA~12.2
  \item {\bf Model:} fine-tuned LLaMA-3.1-8B
  \item {\bf Data set:} retain/forget datasets; links to sensitive models
  \item {\bf Hardware:} 1--2$\times$A100 (40/80\,GB)
  \item {\bf Execution:} \texttt{salad\_unlearn.sh}, \texttt{salad\_eval.sh}
  \item {\bf Metrics:} FR, FP, Min-K\%, Min-K\%++, PrivLeak; utility (pass@K)
  \item {\bf Output:} generated Verilog; evaluation logs
  \item {\bf Experiments:} 4 threat scenarios $\times$ 6 methods $\times$ 3 epochs
  \item {\bf Publicly available?:} \href{https://github.com/DfX-NYUAD/SALAD}{github.com/DfX-NYUAD/SALAD}
  \item {\bf Workflow framework used?:} TOFU OpenUnlearning~\cite{maini2024tofu}
\end{itemize}
}

\subsection{Description}

\subsubsection{How to Access}
\href{https://github.com/DfX-NYUAD/SALAD}{github.com/DfX-NYUAD/SALAD}




\subsubsection{Datasets}

\noindent\textbf{Retain dataset.} The retain dataset consists of 27K samples from the RTL-Coder training dataset~\cite{RTLCoder}.
\textbf{Forget datasets.} Four forget datasets are constructed to align with the threat scenarios:
\begin{enumerate}[label=(\roman*),leftmargin=*]
  \item \textit{Benchmark contamination:} 156 samples from VerilogEval~\cite{liu2023verilogeval} and 50 samples from RTLLM~\cite{lu2024rtllm}.
  \item \textit{Custom designs:} 1,134 filtered design samples from RTL-Repo~\cite{allam2024rtl}.
  \item \textit{Malicious codes:} 1 malicious code example from RTL-Breaker~\cite{mankali2024rtl}.
  \item \textit{IP leakage:} 2 IP leakage examples from VeriLeaky~\cite{wang2025verileaky}.
\end{enumerate}

\subsubsection{Models}
Sensitive model links are provided via the repository (see the \texttt{README} for access instructions)

\subsection{Installation}

Please refer to the repository \texttt{README} for environment setup and installation instructions.

\subsection{Experiment workflow}
Run \texttt{salad\_unlearn.sh} on the provided sensitive models to produce the unlearned model. Then run \texttt{salad\_eval.sh} to evaluate unlearning performance; results are written to the logs.
\subsection{Evaluation and expected results}
The corresponding evaluation results (FR and Min-K\%++) are shown in Figs.~\ref{fig:VerilogEval_unlearn_exp}, \ref{fig:plot_customip}, \ref{fig:malicious_exp}, and \ref{fig:leaky_exp}; additional results (FP, Min-K\%, and PrivLeak) are provided in Fig.~\ref{fig:results_ext}. Generated Verilog outputs are written to the evaluation logs.

\end{document}